\documentclass{article}
\usepackage{spconf,amsmath,graphicx}
\usepackage{spconf,amsmath,graphicx}
\usepackage{hyperref}
\usepackage{times}
\usepackage{epsfig}
\usepackage{amssymb}
\usepackage{booktabs}
\usepackage{stix}
\usepackage{algpseudocode}
\usepackage{algorithm} 
\usepackage{subcaption}
\usepackage{caption}
\usepackage{multicol}
\usepackage{tabularx}
\usepackage{multirow}

\title{PAS-MEF: Multi-Exposure Image Fusion based on Principal Component Analysis, Adaptive Well-Exposedness and Saliency Map}
\name{Diclehan Karakaya, Oguzhan Ulucan, Mehmet Turkan}
\address{Department of Electrical and Electronics Engineering, Izmir University of Economics, Izmir, Turkey}

\begin{document}
\ninept
\maketitle
\begin{abstract}
	High dynamic range (HDR) imaging enables to immortalize natural scenes similar to the way that they are perceived by human observers. With regular low dynamic range (LDR) capture/display devices, significant details may not be preserved in images due to the huge dynamic range of natural scenes. To minimize the information loss and produce high quality HDR-like images for LDR screens, this study proposes an efficient multi-exposure fusion (MEF) approach with a simple yet effective weight extraction method relying on principal component analysis, adaptive well-exposedness and saliency maps. These weight maps are later refined through a guided filter and the fusion is carried out by employing a pyramidal decomposition. Experimental comparisons with existing techniques demonstrate that the proposed method produces very strong statistical and visual results.
\end{abstract}


\section{Introduction}
\label{sec:intro}

	High dynamic range (HDR) technology aims at producing high quality images similar to human perception. However, the dynamic range gap between high contrast scenes and low dynamic range (LDR) capture/display devices causes information loss in highlights and shadows~\cite{mertens2009exposure}. In order to minimize distortions and detail loss, i.e., to capture and display high quality images, there are three main approaches: \textit{(i)} using HDR compatible capture and display devices, \textit{(ii)} employing tone-mapping operators to map HDR onto LDR displays, and \textit{(iii)} using multi-exposure LDR fusion (MEF) to create HDR-like content for LDR screens~\cite{ulucan2020multi}. The user-grade technology manufacturers generally prefer to use MEF to obtain high quality LDR images, since MEF has significantly lower cost than hardware-based solutions; and with MEF, it is also possible to avoid tone-mapping related problems such as low-subjective contrast and color saturation~\cite{akyuz2006color,kiser2012real}.
	
	MEF mainly aims at keeping the most informative parts of each exposure image via extracting weight maps, and then it blends them into a single HDR-like image~\cite{mertens2009exposure}. There are several MEF studies present in the literature. In the milestone study of Mertens~\textit{et~al.}~\cite{mertens2009exposure}, a weight map extraction scheme is proposed which is based on contrast, saturation and well-exposedness. The fusion of the input stack is carried out by taking the Gaussian pyramid of weight maps and the Laplacian pyramid of exposures, which is inspired from Burt and Kolczynski~\cite{burt1993enhanced}. In a recent study of Lee~\textit{et~al.}~\cite{lee2018multi}, a weight map is formed by employing an adapted version of well-exposedness of Mertens and a second map is characterized via the gradient information in each exposure. Finally, a pyramidal image decomposition is employed to carry out the fusion process. In Li and Zhang~\cite{li2018multi} (Li$18$), convolutional neural networks (CNNs) are employed for MEF. The first layer of a pretrained classification network is used for feature extraction. Then, local visibility and temporal consistency maps are extracted and adopted for the weighted fusion operation. In Liu and Leung~\cite{liu2019variable}, a method is proposed for both MEF and decolorization. In this study, the local gradient information of each exposure is extracted and provided to a CNNs model. The proposed algorithm produces satisfying results, however it can operate on stacks consisting of three exposures only. In Hayat and Imran~\cite{hayat2019ghost}, the weight map characterization scheme of Li and Kang~\cite{ShutaoLi} is modified with the dense-SIFT descriptor, and guided filtering is used to eliminate discontinuities and noise in these maps. Finally, image fusion is conducted via a pyramid decomposition approach. In the study of Li~\textit{et~al.}~\cite{li2020fast} (Li$20$), the method of Ma~\textit{et al.}~\cite{ma2017robust} is investigated and improved by forming adopted weight maps via signal strength, signal structure and mean intensities. Recursive downsampling and processing are included into the model and halo effects are significantly reduced. In a recent study of Ulucan~\textit{et al.}~\cite{ulucan2020multi}, a MEF algorithm is developed which is based on linear embeddings of images and watershed masking. Weights maps are first characterized via linear embeddings of exposure image patch spaces, while preserving local geometry of the sampled manifold structure. These weight maps are then refined via watershed masking to highlight most informative parts of each exposure in the input stack. Lastly, the fusion process is performed through weighted averaging.
	
	As it can be deduced from above studies, each specific algorithm generally differs in the way of extraction and/or characterization of weight maps. Therefore, new weight map extraction methods will enlighten the path leading to a general map formation framework. To this end, in this paper, a novel MEF method is proposed to fuse static exposure stacks. The weight map extraction algorithm relies on principal component analysis (PCA), adaptive well-exposedness and saliency map features. These maps are later refined by a guided filter, and then exposure images are fused via a pyramidal decomposition. The proposed method is compared with well-known MEF algorithms and it demonstrates very strong outputs both statistically and visually. To the best of available knowledge, this is the first time that PCA is used for extracting weights in MEF. Furthermore, this is the first study that modifies the well-exposedness (i.e., indicating brightness) feature to be fully adaptive, while existing MEF algorithms employ a fixed parameter and/or constant for well-exposedness. Finally, it is worth mentioning here that saliency maps are used in this study to mimic the human visual system (HVS) and give larger weights to the ``best'' parts of images as in the primary visual cortex.
	
	This paper is organized as follows. The proposed MEF technique is detailed in Sec.~\ref{sec:method}. Experimental results are presented and discussed in Sec.~\ref{sec:expres}. Lastly, a brief summary and possible future directions for this study are given in Sec.~\ref{sec:conc}.

\section{The Proposed MEF Method: PAS-MEF}
\label{sec:method}

\begin{figure}[!t]
    \centering
    \includegraphics[width=0.48\textwidth]{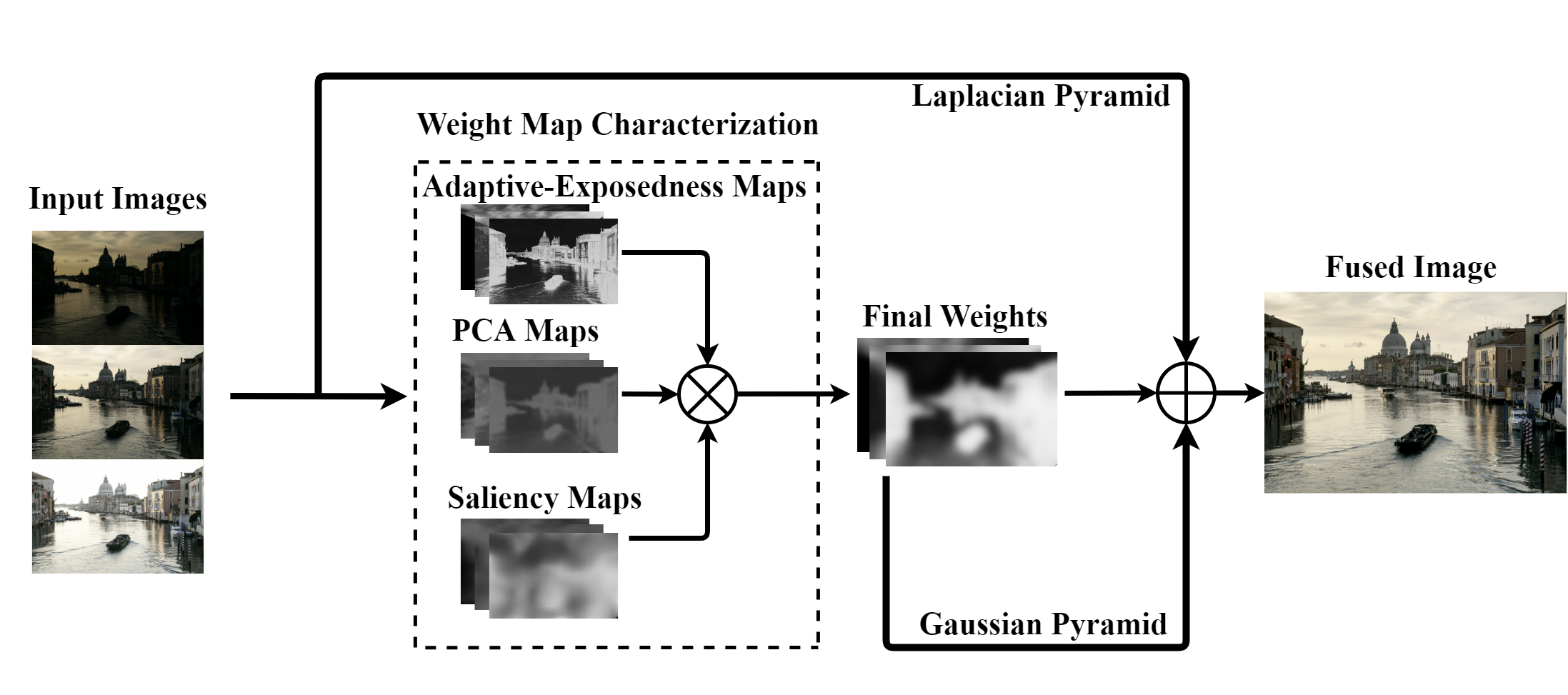}
    \caption{A flowchart of the proposed MEF method.}
    \label{fig:flowchart}
\end{figure}

A simple flowchart of the proposed method is given in Fig.~\ref{fig:flowchart}. Given the input exposures, there are three branches for extracting PCA, adaptive well-exposedness and saliency maps. These maps are later combined to obtain final fusion weights in order to output a fused image via a pyramidal decomposition.

{\bf Weight extraction via PCA.~}~PCA performs an orthogonal transformation to obtain linearly uncorrelated variables from possibly correlated data~\cite{wold1987principal}. The correlated data is projected onto the PCA space by taking advantage of the eigenvectors of the covariance matrix. The first principal component is in the direction of the highest variance (first eigenvector), while the second principal component lies in the subspace perpendicular to the first one and the next principal components are computed similarly. The representations of data in the PCA space can be called as \textit{scores}.

\begin{figure}[!t]
	\centering
	\includegraphics[width=0.155\textwidth]{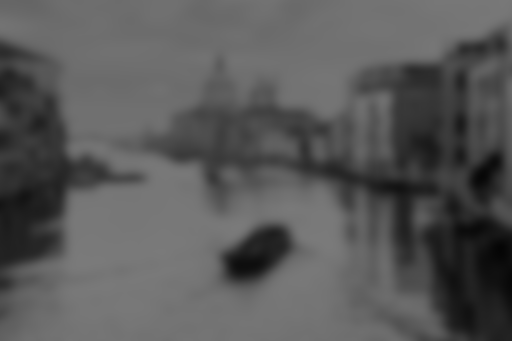}
	\includegraphics[width=0.155\textwidth]{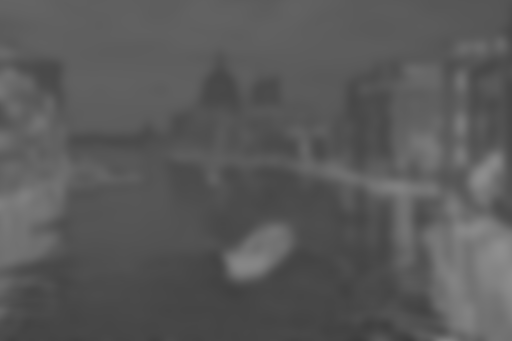}
	\includegraphics[width=0.155\textwidth]{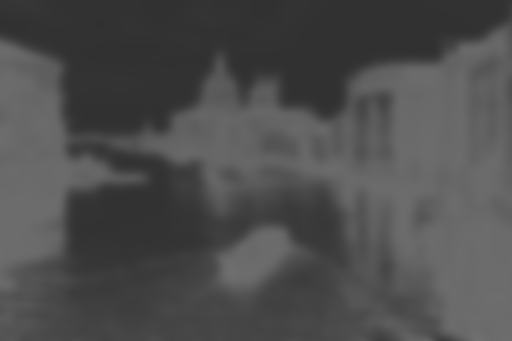}
	\includegraphics[width=0.155\textwidth]{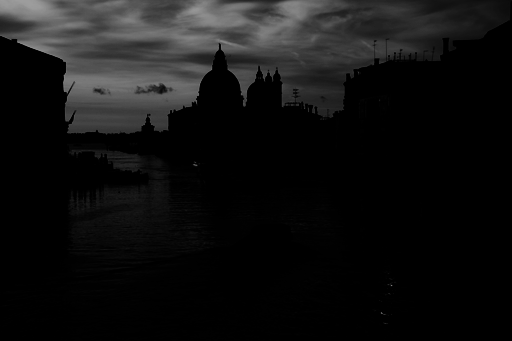}
	\includegraphics[width=0.155\textwidth]{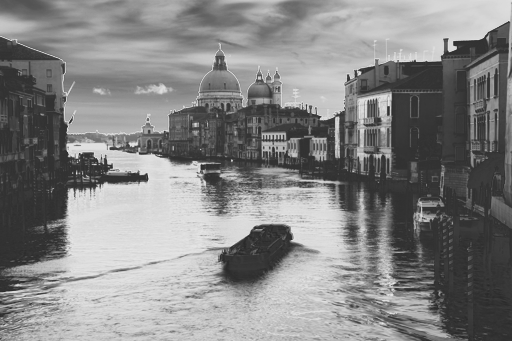}
	\includegraphics[width=0.155\textwidth]{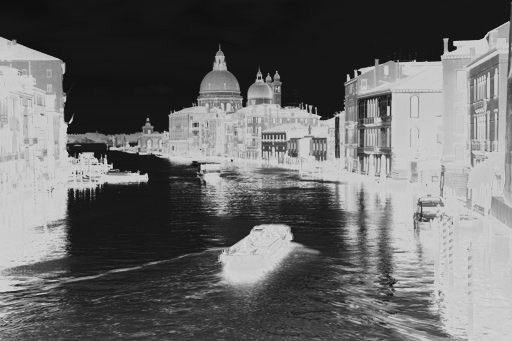}
	\includegraphics[width=0.155\textwidth]{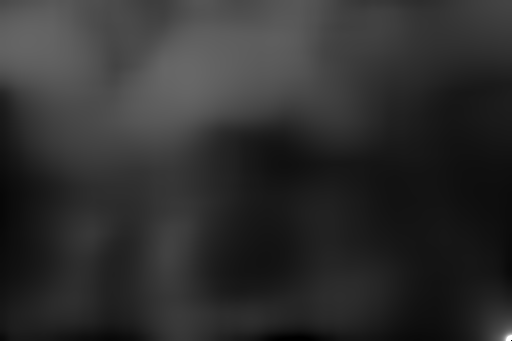}
	\includegraphics[width=0.155\textwidth]{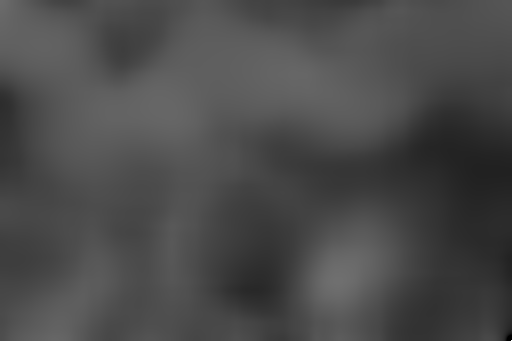}
	\includegraphics[width=0.155\textwidth]{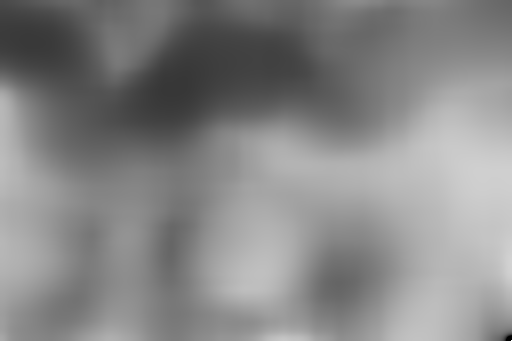}
	\includegraphics[width=0.155\textwidth]{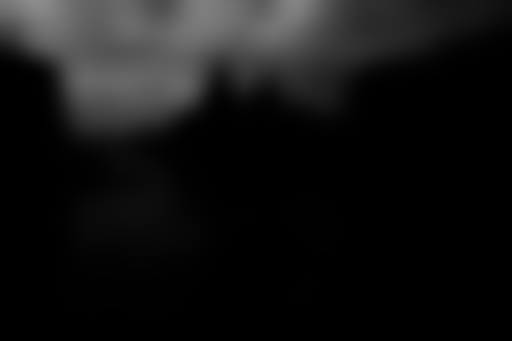}
	\includegraphics[width=0.155\textwidth]{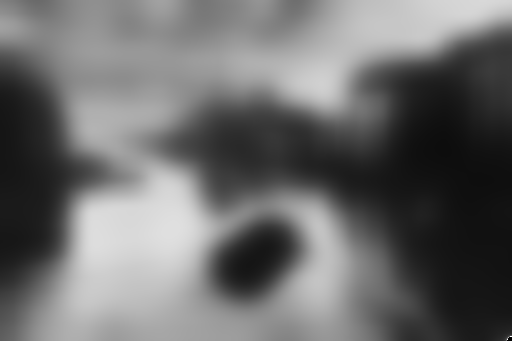}
	\includegraphics[width=0.155\textwidth]{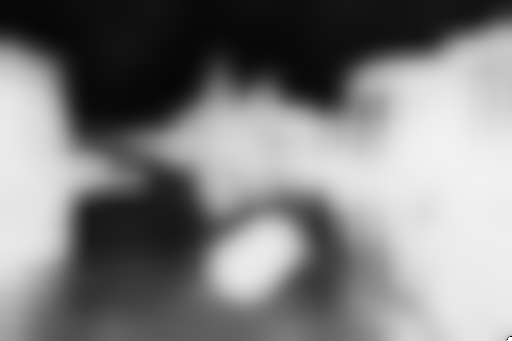}
	\caption{Weights for \textit{Venice}. (Top-to-bottom) PCA maps, adaptive well-exposedness weights, saliency maps, final fusion weights.}
	\label{fig:weights}
\end{figure}

PCA has already been used in image fusion but, to the best of available knowledge, it has not been employed in MEF studies~\cite{li2017pixel}. Therefore, it is investigated in this study first by vectorizing $N$ number of gray-scale versions of exposure images $I_n, n=1\dots N,$ into column vectors of the size $rc \times 1$ where $r$ and $c$ represent the number of rows and columns of each image, respectively. The obtained $N$ column vectors are then stacked into the columns of an $rc \times N$ data matrix, in which there are $rc$ observations with $N$ variables each, for calculating the scores of observations via PCA. Subsequently, each variable-score vector is linearly normalized to have a range between $[0~1]$, then reshaped back to an $r\times c$ matrix followed by a simple smoothing Gaussian filter. Finally, $N$ number of PCA weight maps ($P_n$) are obtained with a sum-to-one normalization applied at each spatial position. As an example, the extracted PCA maps of the \textit{Venice} stack are demonstrated in Fig.~\ref{fig:weights}.

{\bf Weight extraction via adaptive well-exposedness.~}~The well-exposedness feature is initially introduced by Mertens. For a given exposure image $I_n$, it is extracted for each red-green-blue channel separately using a Gaussian curve as $exp\left(-( I_n-0.5)^2/2\sigma^2\right)$ where $\sigma = 0.2$. This weight aims at keeping pixel intensities which are not too close to $0$ (under-exposed) or $1$ (over-exposed), hence it favors pixels in well-exposed regions with intensity values close to $0.5$. However, this exposedness feature sometimes can not sufficiently preserve bright regions of short-exposure images, as well as dark regions of long-exposure images~\cite{lee2018multi}. To overcome this problem, Lee proposed an adaptive well-exposedness based on the mean of pixel intensities of exposure images. In this adaptive form, the constant values of $0.5$ and $\sigma$ of Mertens are replaced with some functions of the mean of pixel intensities of $I_n$ and its neighboring exposure images $I_{n-1}$ and $I_{n+1}$. However, there still exists a fixed constant parameter to calculate the $\sigma$ value.

In this study, a fully adaptive weight calculation is proposed for the well-exposedness scheme in order to allocate large weights for dark regions when the image is a long-exposure, as well as for bright regions when the image is a short-exposure. All computations are carried out on the luminance channel $Y_n$ of $I_n$ as given in Eqn.~(\ref{eq:adwell}),
\begin{equation}
	\footnotesize
	\centering
	A_n = exp \left( -\frac{\left( Y_n-(1-\mu_{Y_n}) \right)^2}{2\sigma_{Y_n}^2} \right) 
	\label{eq:adwell}
\end{equation}
where $\mu_{Y_n}$ and $\sigma_{Y_n}$ represent the mean of pixel intensities and the standard deviation of $Y_n$, respectively. This finally leads to an adaptive algorithm since the Gaussian curve controlling parameters are extracted via self statistical information of each single exposure, and larger weights are given to the ``best'' luminance intensities of the input. The obtained adaptive well-exposedness weights ($A_n$) of the \textit{Venice} stack are illustrated in Fig.~\ref{fig:weights}.

{\bf Weight extraction via saliency map.~}~When processed in the HVS, the attention that objects gather depends on the task at hand and stimulus-driven factors such as prominent colors~\cite{bruce2009saliency}. In the literature, several computational models are proposed to mimic the HVS; thus to highlight salient regions, increase the visual appeal and quality of images.

In this study, saliency maps are used for assigning larger weights to regions which are more attractive to human observers. Since the design of a salient region detection algorithm is out of context, the technique introduced by Hou~\textit{et al.}~\cite{hou2011image} is integrated into the proposed MEF method. This saliency algorithm is based on a descriptor called \textit{image signature}, which is the sign of the Discrete Cosine Transform (DCT) coefficients. Briefly, the DCT is first computed for each red-green-blue channel separately. Then, image reconstruction is carried out by calculating the inverse DCT of the sign of the DCT coefficients. The saliency maps ($S_n$) are finally obtained from the reconstructed image. For detailed information, the reader may refer to~\cite{hou2011image}. The saliency maps of the \textit{Venice} stack are shown in Fig.~\ref{fig:weights}.

{\bf Weight refinement and fusion.~}~After all three weight maps are characterized, they are combined to form a single refined map for each exposure in the input stack given in Eqn.~(\ref{weightcomb}) as follows,
\begin{equation}
    W_n = \texttt{GuidFilt}(P_n \times A_n \times S_n),~n=1\dots N,
    \label{weightcomb}
\end{equation}
where \texttt{GuidFilt} is an edge-aware (edge-preserving) smoothing filter called \textit{guided filter}~\cite{he2010guided}, which is generally used to eliminate possible discontinuities and noise in weight maps, e.g.,~\cite{hayat2019ghost}. All these maps $W_n$ are finally normalized to satisfy a sum-to-one constraint at each spatial position to form the final weights for fusion. The obtained final fusion weights of the \textit{Venice} stack are given in Fig.~\ref{fig:weights}.

A pyramidal decomposition is applied to blend the input stack in order to further avoid artifacts such as halo effects at sharp texture and color changeovers~\cite{burt1993enhanced,mertens2009exposure}. In detail, the Laplacian pyramid ($L$) is employed to decompose each input exposure into $\ell$-levels of distinct resolutions, while the Gaussian pyramid ($G$) to carry out a similar operation for final fusion weights. The blending operation is applied at each pyramidal level, and as a result a fused Laplacian pyramid is obtained for the fused image given in Eqn.~(\ref{pyramid}) as follows,
\begin{equation}
	L\{F^\ell\} = \sum_{n=1}^N G\{W_n^\ell\} \times L\{I_n^\ell\}
	\label{pyramid}
\end{equation}
where the fused pyramid $L\{F^\ell\}$ is finally collapsed to acquire the final fused image $F$.
\begin{figure}[!t]
\centering
\includegraphics[width=0.475\linewidth]{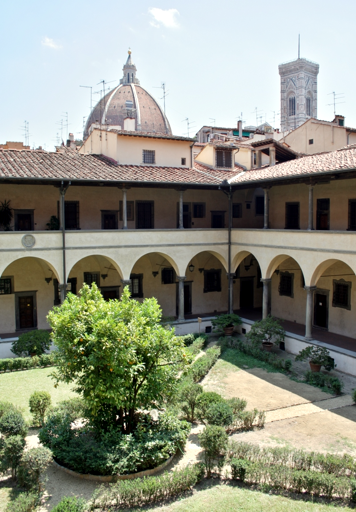}
\includegraphics[width=0.475\linewidth]{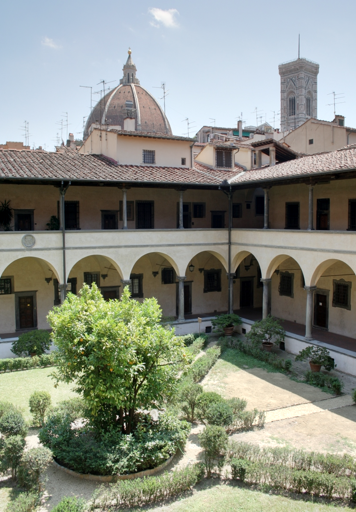}
\caption{\textit{Laurenziana}: (Left) Ulucan ($0.989$), (right) Proposed ($0.991$).} 
\label{fig:Laurenziana}
\end{figure}

\section{Experimental Results}
\label{sec:expres}

The proposed MEF algorithm (PAS-MEF) is compared to Mertens~\cite{mertens2009exposure}, Lee~\cite{lee2018multi}, Li$18$~\cite{li2018multi}, Liu~\cite{liu2019variable}, Hayat~\cite{hayat2019ghost}, Ulucan~\cite{ulucan2020multi} and Li$20$~\cite{li2020fast} over $13$ image stacks obtained from datasets in~\cite{Data1,Data2,merianos2019multiple}. All experiments are performed on an AMD Ryzen(TM) $5$ $3600$x CPU @ $3.80$ GHz $6$-core $16$GB RAM machine using MATLAB R$2020$a. All competing algorithms are employed with their default settings, including the proposed method, without any optimization. A statistical performance analysis is performed through the multi-scale structural similarity framework for MEF, i.e., MEF-SSIM~\cite{Data2}. MEF-SSIM is a perceptual quality assessment metric, which takes both the global luminance consistency and the local structure preservation into account to produce statistical results in the range $[0~1]$. A score closer to $1$ indicates a better perceptual quality.

The obtained statistical scores of all algorithms are reported in Table~\ref{ssimtable}, in which the bottom three rows present the average accuracy (avg), standard deviation (std) and average  execution time (run-time) of each method. It can be clearly observed that PAS-MEF produces highly competitive results (i.e., being in the second best place) and surpasses six of the competing state-of-the-art MEF approaches on average. In addition, these statistical results indicate that the standard deviation of PAS-MEF scores is the smallest (together with Li$20$) and the computational complexity is very conceivable.
\begin{table*}[!t]
\centering
\footnotesize
	\caption{MEF-SSIM scores for each exposure stack used in experiments. The highest scores are in boldface.}
\begin{tabular}{lccccccc|c}
    \toprule
                        & Mertens   & Lee      & Li$18$    & Liu     & Hayat   & Ulucan     & Li$20$    & PAS-MEF \\ \hline
\textit{Arno}           & $\mathbf{0.991}$   & $0.987$  & $0.948$ & $0.985$ & $0.985$ & $0.986$  & $0.990$ & $0.989$  \\ \hline
\textit{Chinese Garden} & $0.989$   & $0.990$  & $0.977$ & $0.988$ & $0.993$ & $0.991$  & $\mathbf{0.994}$ & $0.993$  \\ \hline
\textit{Church}         & $0.989$   & $\mathbf{0.992}$  & $0.980$ & $0.977$ & $\mathbf{0.992}$ & $0.989$  & $\mathbf{0.992}$ & $0.991$  \\ \hline
\textit{Farmhouse}      & $0.981$   & $0.979$  & $0.974$ & $0.978$ & $0.984$ & $0.978$  & $\mathbf{0.986}$ & $0.981$  \\ \hline
\textit{Flowers}        & $0.964$   & $0.990$  & $0.972$ & $0.990$ & $\mathbf{0.995}$ & $0.989$  & $\mathbf{0.995}$ & $0.990$  \\ \hline
\textit{Kluki}          & $0.980$   & $0.974$  & $0.957$ & $0.973$ & $0.980$ & $0.963$  & $\mathbf{0.983}$ & $0.979$  \\ \hline
\textit{Laurenziana}    & $0.988$   & $0.987$  & $0.973$ & $0.987$ & $0.989$ & $0.989$  & $0.989$ & $\mathbf{0.991}$  \\ \hline
\textit{Lighthouse}     & $0.980$   & $0.979$  & $0.962$ & $\mathbf{0.985}$ & $0.974$ & $0.975$  & $0.978$ & $0.982$  \\ \hline
\textit{Mask}           & $0.987$   & $0.990$  & $0.975$ & $0.985$ & $\mathbf{0.992}$ & $0.987$  & $0.991$ & $\mathbf{0.992}$  \\ \hline
\textit{Office}         & $0.985$   & $\mathbf{0.991}$  & $0.970$ & $0.985$ & $0.987$ & $\mathbf{0.991}$  & $0.989$ & $0.984$  \\ \hline 
\textit{OldHouse}       & $0.974$   & $0.990$  & $0.962$ & $0.988$ & $0.968$ & $0.991$  & $0.990$ & $\mathbf{0.992}$  \\ \hline 
\textit{Tower}          & $0.986$   & $0.987$  & $0.981$ & $0.983$ & $0.987$ & $0.982$  & $\mathbf{0.988}$ & $0.983$  \\ \hline 
\textit{Venice}         & $0.966$   & $0.972$  & $0.947$ & $0.973$ & $0.972$ & $0.978$  & $\mathbf{0.984}$ & $0.980$  \\ \hline \hline
\textbf{avg}        & $0.981$   & $0.985$  & $0.967$ & $0.982$ & $0.984$ & $0.983$  & $\mathbf{0.988}$ & $0.986$  \\ 
\textbf{std}            & $0.009$  & $0.007$ & $0.011$ & $0.006$ & $0.009$ & $0.008$  & $\mathbf{0.005}$ & $\mathbf{0.005}$  \\ \hline \hline 
\textbf{run-time (sec)}   & $0.36$    & $0.38$   & $0.93$  & $50.68$ & $0.89$  & $2.91$   & $\mathbf{0.31}$  & $0.66$    \\ \bottomrule
\label{ssimtable}
\end{tabular}
\end{table*}
\begin{figure}[!t]
\centering
\includegraphics[width=0.4525\linewidth]{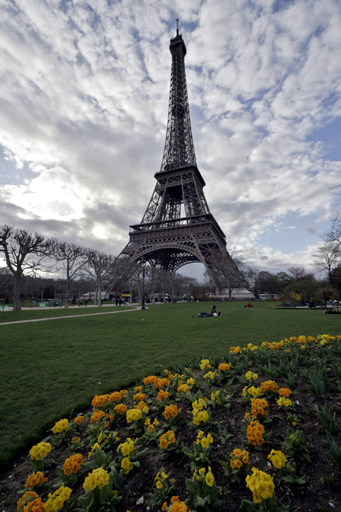}
\includegraphics[width=0.4525\linewidth]{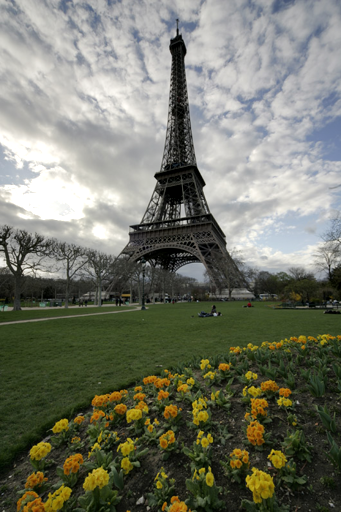}
\caption{\textit{Tower}: (Left) Ulucan ($0.982$), (right) Proposed ($0.983$).} 
\label{fig:tower}
\end{figure}

A side-by-side visual comparison of the fusion outputs of PAS-MEF and Ulucan is given for the \textit{Laurenziana} stack in Fig.~\ref{fig:Laurenziana}. In this particular example, PAS-MEF reaches the highest MEF-SSIM score when compared to other competing techniques. It can be clearly seen that the sky has a more natural color in the output of PAS-MEF. Furthermore, the details on the rooftops are better preserved in the result of this technique, while the tree in the middle has more vivid colors in Ulucan. Another visual comparison for the \textit{Tower} stack is demonstrated in Fig.~\ref{fig:tower}. The grass and flowers on the foreground are much better recovered in the proposed PAS-MEF, while clouds are better preserved and the tower has vivid colors in Ulucan. 

\begin{figure}[!t]
\centering
\includegraphics[width=0.475\linewidth]{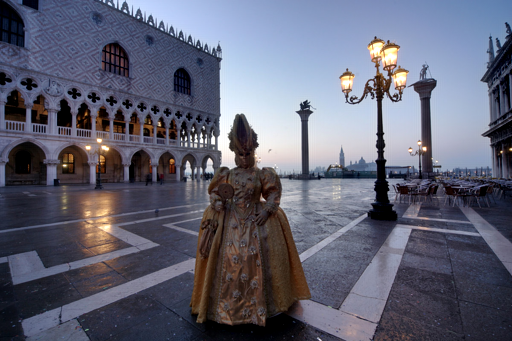}
\includegraphics[width=0.475\linewidth]{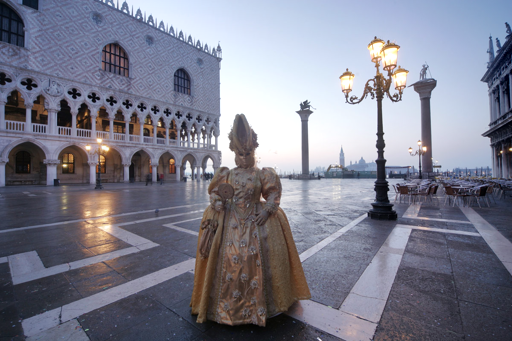}
\caption{\textit{Mask}: (Left) Li$20$ ($0.991$), (right) Proposed ($0.992$).} 
\label{fig:Mask}
\end{figure}

In Fig.~\ref{fig:Mask}, the fusion results are presented for the \textit{Mask} stack. PAS-MEF produces the highest MEF-SSIM score (together with Hayat) for this exposure sequence. When compared to Li$20$ which has lower brightness and less information in several parts in the fused image, the proposed method clearly preserves the details of the building and the mask. On the other side, the sky has a more plausible color in Li$20$.

\begin{figure}[!t]
\centering
\includegraphics[width=0.475\linewidth]{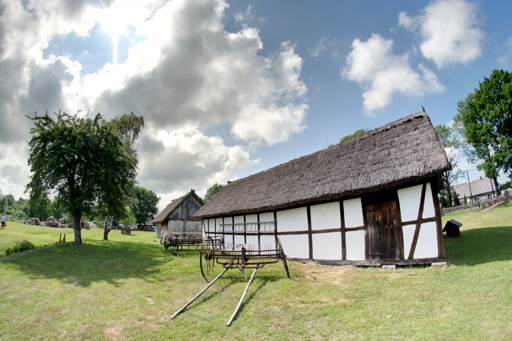}
\includegraphics[width=0.475\linewidth]{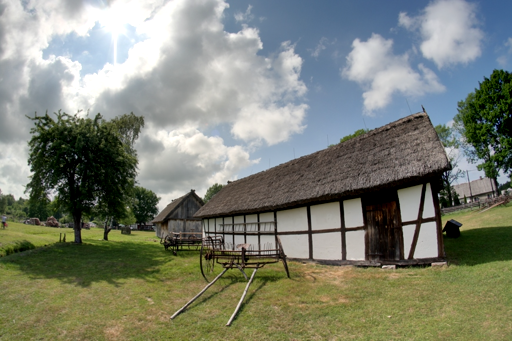}
\caption{\textit{Kluki}: (Left) Hayat ($0.980$), (right) Proposed ($0.979$).} 
\label{fig:Kluki}
\end{figure}

In Fig.~\ref{fig:Kluki}, the fusion outputs of PAS-MEF and Hayat are compared for the \textit{Kluki} stack. Although PAS-MEF has its lowest MEF-SSIM score for this exposure sequence among other stacks in the dataset, the sky has more well-settled colors in blue regions, and the rooftop of the house and the grass have more vivid colors when compared to Hayat, whose MEF-SSIM is slightly higher.

\begin{figure}[!t]
\centering
\includegraphics[width=0.475\linewidth]{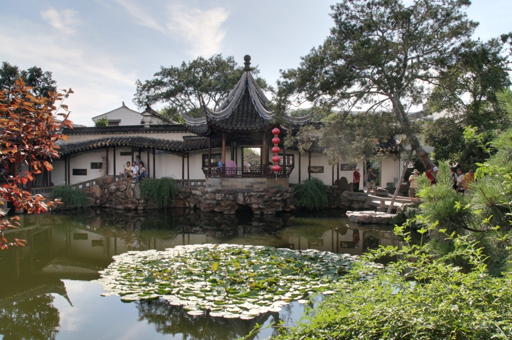}
\includegraphics[width=0.475\linewidth]{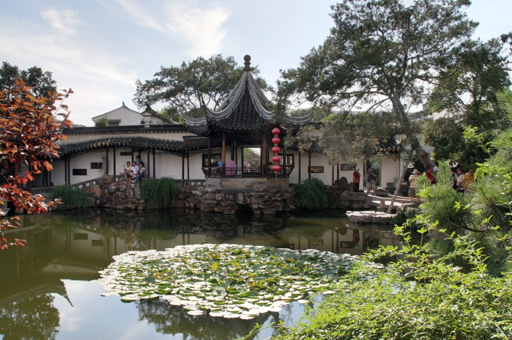}
\caption{\textit{Chinese Garden}: (Left) Lee ($0.990$), (right) Proposed ($0.993$).} 
\label{fig:chinese}
\end{figure}

\begin{figure}[!t]
\centering
\includegraphics[width=0.475\linewidth]{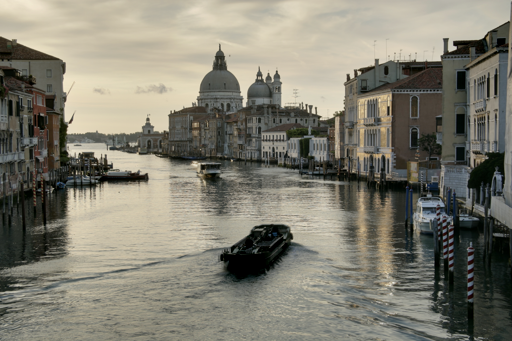}
\includegraphics[width=0.475\linewidth]{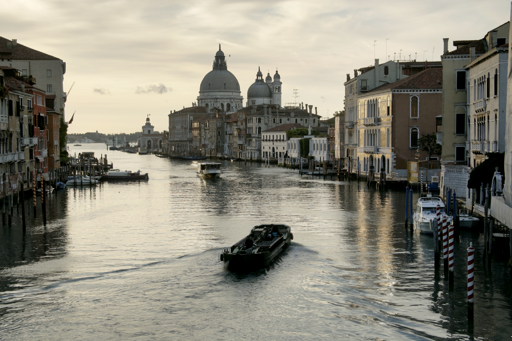}
\caption{\textit{Venice}: (Left) Lee ($0.972$), (right) Proposed ($0.980$).} 
\label{fig:Venice}
\end{figure}

It can be concluded from Table~\ref{ssimtable} that PAS-MEF reaches its highest MEF-SSIM score for the \textit{Chinese Garden} stack and some visual fusion results are presented in Fig.~\ref{fig:chinese}. When compared to Lee, the sky region is more plausible in the proposed technique and overall a natural-looking image is obtained while avoiding any artifacts. In addition, it can be observed from Fig.~\ref{fig:Venice} that both PAS-MEF and Lee output natural-looking images for \textit{Venice}. However, the sky and buildings on the left contain more vivid colors in PAS-MEF, which led to a significantly higher MEF-SSIM score than Lee. 

\begin{figure}[!t]
\centering
\includegraphics[width=0.475\linewidth]{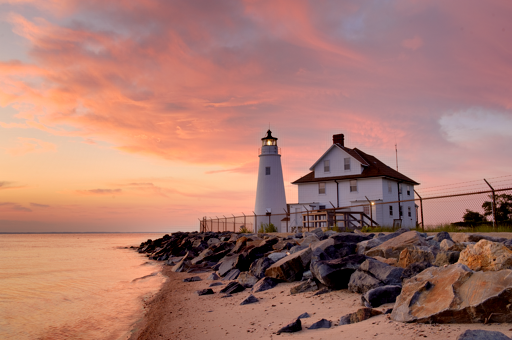}
\includegraphics[width=0.475\linewidth]{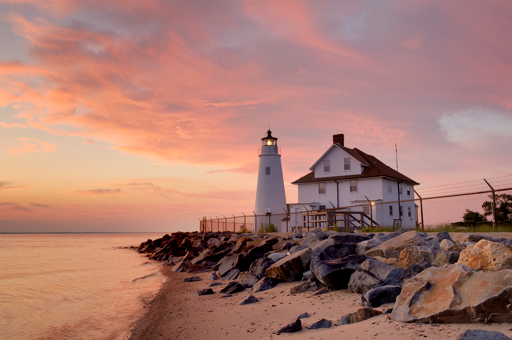}
\caption{\textit{Lighthouse}: (Left) Li$20$ ($0.978$), (right) Proposed ($0.982$).} 
\label{fig:Lighthouse}
\end{figure}

\begin{figure}[!t]
\centering
\includegraphics[width=0.475\linewidth]{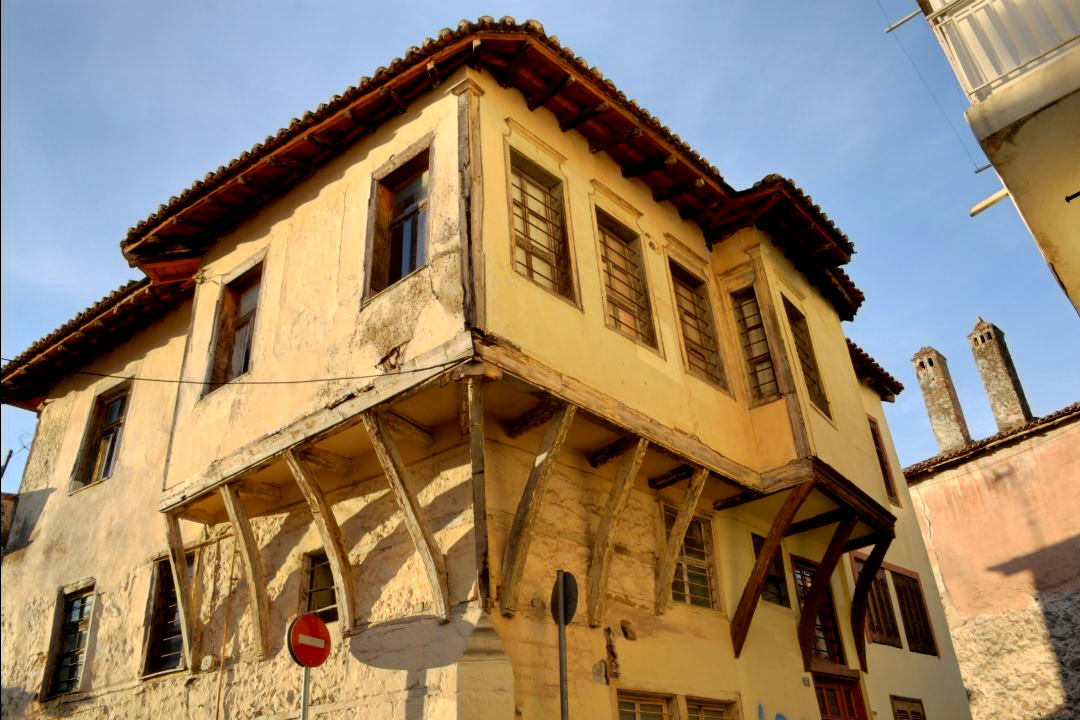}
\includegraphics[width=0.475\linewidth]{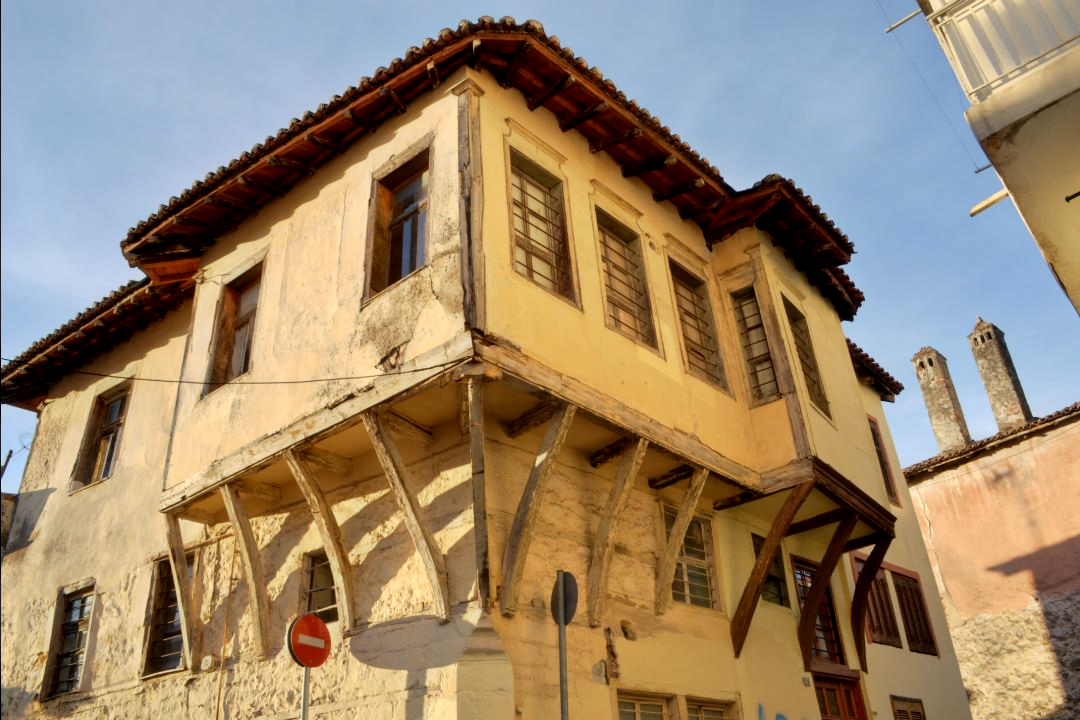}
\caption{\textit{OldHouse}: (Left) Li$20$ ($0.990$), (right) Proposed ($0.992$).} 
\label{fig:OldHouse}
\end{figure}

Further fusion examples are illustrated in Fig.~\ref{fig:Lighthouse} and Fig.~\ref{fig:OldHouse} for \textit{Lighthouse} and \textit{OldHouse}, respectively. The rooftop of the light house, rocks and trees in the background have more striking colors in PAS-MEF, while a darker output is generated by Li$20$. Next in Fig.~\ref{fig:OldHouse}, Li$20$ contains very bright regions on the old house. The building has more well-settled colors in PAS-MEF which has the highest MEF-SSIM score when compared to other methods.

\section{Conclusion}
\label{sec:conc}
MEF is commonly used for obtaining HDR-like high quality images and numerous studies are present in this field. In general, the existing methods differ in the weight map characterization process. In this study, a novel weight extraction method is introduced which is based on PCA, adaptive well-exposedness and saliency maps. The obtained weights are refined via a guided filter and then image fusion is carried out through a pyramidal decomposition. The proposed algorithm presents very strong results both statistically and visually, and it outperforms several state-of-the-art MEF techniques. It is worth noting here that, to the best of available knowledge, this is the first study which incorporates PCA and fully adaptive well-exposedness into the MEF problem.

As a future direction, the proposed algorithm will be optimized for increasing its statistical and visual performance, and for further reducing the run-time complexity. Moreover, PAS-MEF will be extended to fuse dynamic scenes in the image stack.

\bibliographystyle{IEEE.bst}
\bibliography{refs.bib}

\end{document}